\documentstyle[aaai]{article}
\input{psfig}

\title{Multi-document Summarization by Graph Search and Matching}
\twocolumn

\author{Inderjeet Mani \\
The MITRE Corporation \\
7525 Colshire Drive, W640 \\
McLean, VA 22102, USA \\
{\tt imani@mitre.org} \And
Eric Bloedorn \\
The MITRE Corporation \\
7525 Colshire Drive, W640 \\
McLean, VA 22102, USA \\
{\tt  bloedorn@mitre.org}}

\begin{document}
\bibliographystyle{named}

\maketitle

\begin{abstract}
We describe a new method for summarizing similarities and differences
in a pair of related documents using a graph representation for text.
Concepts denoted by words, phrases, and proper names in the document are
represented positionally as nodes in the graph along with edges
corresponding to semantic relations between items.  Given a
perspective in terms of which the pair of documents is to be
summarized, the algorithm first uses a spreading activation technique
to discover, in each document, nodes semantically related to the
topic. The activated graphs of each document are then matched to yield
a graph corresponding to similarities and differences between the
pair, which is rendered in natural language.  An evaluation of these
techniques has been carried out.
\end{abstract}


\section{Introduction\footnotemark}

\footnotetext{Copyright \copyright 1997, American Association for
Artificial Intelligence ({\tt www.aaai.org}). All rights
reserved.}

With the mushrooming of the quantity of on-line text information,
triggered in part by the growth of the World Wide Web, it is
especially useful to have tools which can help users digest
information content. Text summarization attempts to address this
problem by taking a partially-structured source text, extracting
information content from it, and presenting the most important content
to the user in a manner sensitive to the user's needs.  In exploiting
summarization, many modern information retrieval applications need
summarization systems which scale up to large volumes of unrestricted
text.  In such applications, a common problem which arises is the
existence of multiple documents covering similar information, as in
the case of multiple news stories about an event or a sequence of
events. A particular challenge for text summarization is to be able to
summarize the similarities and differences in information {\it
content} among these documents in a way that is sensitive to the needs
of the user.

In order to address this challenge, a suitable representation for
content must be developed. Most fieldable text summarization systems
which aim at scalability (e.g., \cite{EchoSearch}, \cite{Rau},
\cite{Kupiec}, etc.) provide a capability to extract sentences (or
other units) that match the relevance criteria used by the
system. However, they don't attempt to understand the concepts in the
text and their relationships; in short, they don't represent the
meaning of the text. In the ideal case, the meaning of each text would
be made up, say, of the meanings of sentences in the text, which in
turn would be made up of the meanings of words. While the ideal case
is currently infeasible beyond a small fragment of a natural language,
it is possible to arrive at approximate representations of meaning.
In this paper, we propose an approach to scalable text summarization
which builds an abstract content representation based on explicitly
representing entities and the {\it relations} between entities, of the
sort that can be robustly extracted by current information extraction
systems. Here, concepts described in a document (denoted by text items
such as words, phrases, and proper names) are represented positionally
as nodes in a graph along with edges corresponding to semantic and
topological relations between concepts. The relations between concepts
are whatever relations can be feasibly extracted in the context of the
scalability requirements of an application: these include
specialization relationships (e.g., which can be extracted based on a
thesaurus), as well as association relationships (such as
relationships between people and organizations, or coreference
relationships between entities). Salient regions of the graph can then
be input to further ``synthesis'' processing to eventually yield
natural language summaries which can in general go well beyond
extracts to abstracts or synopses\footnote{However, the implementation
at the time of writing is confined to extracts.}.

It is also important to note that in computing a salience function for
text items, most fieldable text summarization systems do not typically
deal with the context-sensitive nature of the summarization task. A
user may have an interest in a particular topic, which may make
particular text units more salient. To provide a degree of
context-sensitivity, the summarization algorithm described here takes
a parameter specifying the topic (or perspective) with respect to
which the summary should be generated. This topic represents a set of
entry points (nodes) into the graph. To determine which items are
salient, the graph is searched for nodes semantically related to the
topic, using a spreading activation technique. This approach differs
from other network approaches (such as the use of neural nets, e.g.,
the Hopfield net approach discussed in \cite{Arizona}) in two ways:
first, the structure of our graph reflects both semantic relations
derived from text as well as linear order in the text (the latter via
the positional encoding); the linear order is especially important for
natural language. Second, as will be clarified below, the set of nodes
which become highly activated is a function of link type and distance
from entry nodes, unlike other approaches which use a fixed bound
on the number of nodes or convergence to a stable state.

Of course, if we are able to discover, given a topic and a pair of
related documents, nodes in each document semantically related to the
topic, then these nodes and their relationships can be compared to
establish similarities and differences between the document
pair. Given a pair of related news stories about an event or a
sequence of events, the problem of finding similarities and
differences becomes one of comparing graphs which have been activated
by a common topic.  In practice, candidate common topics can be
selected from the intersection of the activated concepts in each graph
(i.e., which will be denoted by words, phrases, or names). This allows
different summaries to be generated, based on the choice of common
topic. Algorithm FSD-Graphs (Find-Similarities-and-Differences) takes
a pair of such activated graphs and compares them to yield
similarities and differences. The results are then subject to
``synthesis'' processing to yield multi-document summaries.

These graph construction and manipulation techniques are highly
scalable, in that they yield useful summaries in a reasonable time
when applied to large quantities of unrestricted text, of the kind
found on the World Wide Web.  In what follows, we first describe the
graph representation and the tools used to build it, followed by a
description of the graph search and graph matching algorithms. We also
provide an evaluation which assesses the usefulness of a variety of
different graph-based multi-document summarization algorithms.

{\small \begin{figure}
\centerline{\psfig{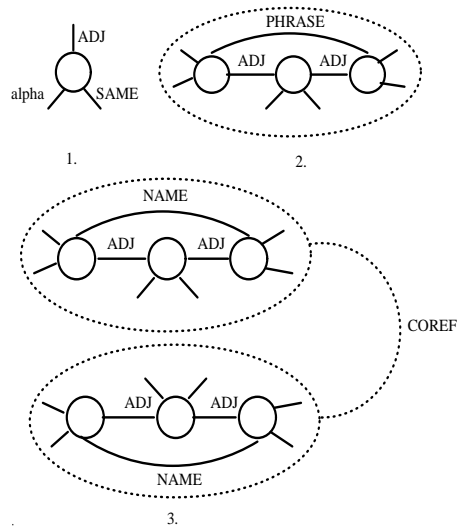}}

\caption{Graph Representation}
\label{fig-1}
\end{figure}}

\section{Representing Meaningful Text Content}

A text is represented as a graph. As shown in Figure~\ref{fig-1}, each
node represents an underlying concept corresponding to a word {\it
occurrence}, and has a distinct input position. Associated with each
such node is a feature vector characterizing the various features of
the word in that position. As shown in part 1 of the figure, a node
can have adjacency links (ADJ) to textually adjacent nodes, SAME links
to other occurrences of the same concept, and other links
corresponding to semantic relationships (represented by $alpha$, to be
discussed below). PHRASE links tie together sequences of adjacent
nodes which belong to a phrase (part 2). In part 3, we show a NAME
link, as well as the COREF link between subgraphs, relating positions
of name occurrences which are coreferential. NAME links can be
specialized to different types, e.g., person, province, etc. The
concepts denoted by phrases and names (indicated by ellipses around
subgraphs in Figure~\ref{fig-1}) are distinguished from the concepts
denoted by words which make up the phrases and names.

\section {Tools for Building Document Graphs}

Our experiments make use of a sentence and paragraph tagger which
contains a very extensive regular-expression-based sentence boundary
disambiguator \cite{Aberdeen}. The boundary disambiguation module is
part of a comprehensive preprocess pipeline which utilizes a list of
75 abbreviations and a series of hand-crafted rules to identify
sentence boundaries. Then, the Alembic part-of-speech tagger
\cite{Aberdeen} is invoked on the text. This tagger uses the rule
sequence learning approach of \cite{Brill}\footnote{When trained on
about 950,000 words of Wall Street Journal text, the tagger obtained
96\% accuracy on a separate test set of 150,000 words of WSJ
\cite{Aberdeen}.}. Names and relationships between names are then
extracted from the document using SRA's NetOwl \cite{Krupka}, a
MUC6-fielded system. Then, salient words and phrases are extracted
from the text using the tf.idf metric, which makes use of a reference
corpus derived from the TREC \cite{Harman} corpus. The weight
$dw_{ik}$ of term $k$ in document $i$ is given by:

{\small \begin{equation}dw_{ik} = tf_{ik} * (log(n)-log(df_{k}) + 1)\end{equation}}
where $tf_{ik}$ = frequency of term k in document i,
$df_{k}$ = number of documents in the reference corpus in which term k occurs, 
$n$ = total number of documents in the reference corpus.

Phrases are useful in summarization as they often often denote
significant concepts, and thus can be good indicators and descriptors
of salient regions of text.  Our phrase extraction method finds
candidate phrases using several patterns defined over part-of-speech
tags. One pattern, for example, uses the maximal sequence of one or
more adjectives followed by one or more nouns. Once stop-words are
filtered out, the weight of a candidate phrase is the average of the
tf.idf weights of remaining (i.e., content) words in the phrase, plus
a factor $\beta$ which adds a small bonus in proportion to the length
of the phrase (to extract more specific phrases). We use a contextual
parameter $\theta$ to avoid redundancy among phrases, by
selecting each term in a phrase at most once. The weight of a phrase W
of length n content words in document i is:

{\small
\begin{equation}wt(W, i) = \beta (n) + \frac{\sum_{k=1}^{n} \theta(ik) * dw_{ik}}{n} \end{equation} }
where $ \theta(ik)$ is 0 if the word has been seen before, and 1 otherwise.

{\small \begin{figure}
\centerline{\psfig{figure=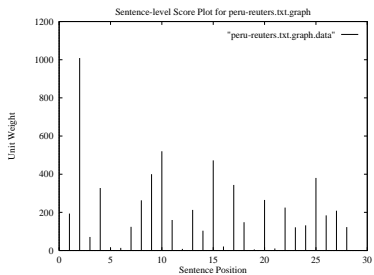,height=1.4in,angle=270}}
\caption{Activation Weights from Raw Graph (Reuters news)}
\label{fig-4a}
\end{figure}}

We now discuss the $alpha$ links. Association relations between
concepts are based on what is provided by NetOwl; for example, {\it
Bill Gates, president of Microsoft} will give rise to the link {\it
president} between the person and the organization.  In lieu of
specialization links between concepts, we initially took the simple
approach of pre-computing the semantic distance links between pairs of
words using Wordnet 1.5 \cite{WordNet}, based on the relative height
of the most specific common ancestor class of the two words, subject
to a context-dependent class-weighting parameter.  For example, for
the texts in Figure~\ref{fig-5}, the words {\it residence} and {\it
house} are very close, because a sense of {\it residence} in WordNet
has {\it house} as an immediate hypernym. This technique is known to
be oversensitive to the structure of the thesaurus. To improve
matters, the corpus-sensitive approach of \cite{Resnick} (see also
\cite{Smeaton}) using the reference corpus has also been implemented;
however, the full exploitation of this, along with suitable
disambiguation techniques will have to await further research.

\section {Graph Search by Spreading Activation}

The goal of the spreading activation algorithm (derived from the
method of \cite{Arizona}) is to find all those nodes that are
semantically linked to the given activated nodes.  The search for
semantically related text is performed by spreading from topic words
to other document nodes via a variety of link types as described
previously. Document nodes whose strings are equivalent to topic
terms (using a stemming procedure $=_{stem}$) are treated as entry
points into the graph. The weight of neighboring nodes is dependent on
the type of node link travelled. For adjacent links, node weight is an
exponentially decaying function of activating node weight and the
distance between nodes. Distances are scaled so that travelling across
sentence boundaries is more expensive than travelling within a
sentence, but less than travelling across paragraph boundaries. For
the other link types, the neighboring weight is calculated as a
function of link weight and activating node weight. The method
iteratively finds neighbors to the given starting nodes (using
$=_{stem}$ in matching strings associated with nodes), pushes the
activating nodes on the output stack and the new nodes on the active
stack and repeats until a system-defined threshold on the number of
output nodes is met, or all nodes have been reached.

As an example, we show the the average weights of nodes at different
sentence positions in the raw graph in Figure~\ref{fig-4a}.  The
results after spreading given the topic {\it Tupac Amaru}, are shown
in Figure~\ref{fig-4b}. The spreading has changed the activation
weight surface, so that some new related peaks have emerged (e.g.,
sentence 4), and old peaks have been reduced (e.g., sentence 2, which
had a high tf.idf score, but was not related to {\it Tupac
Amaru}). The exponential decay function is also evident in the
neighborhoods of the peaks.

{\small \begin{figure}
\centerline{\psfig{figure=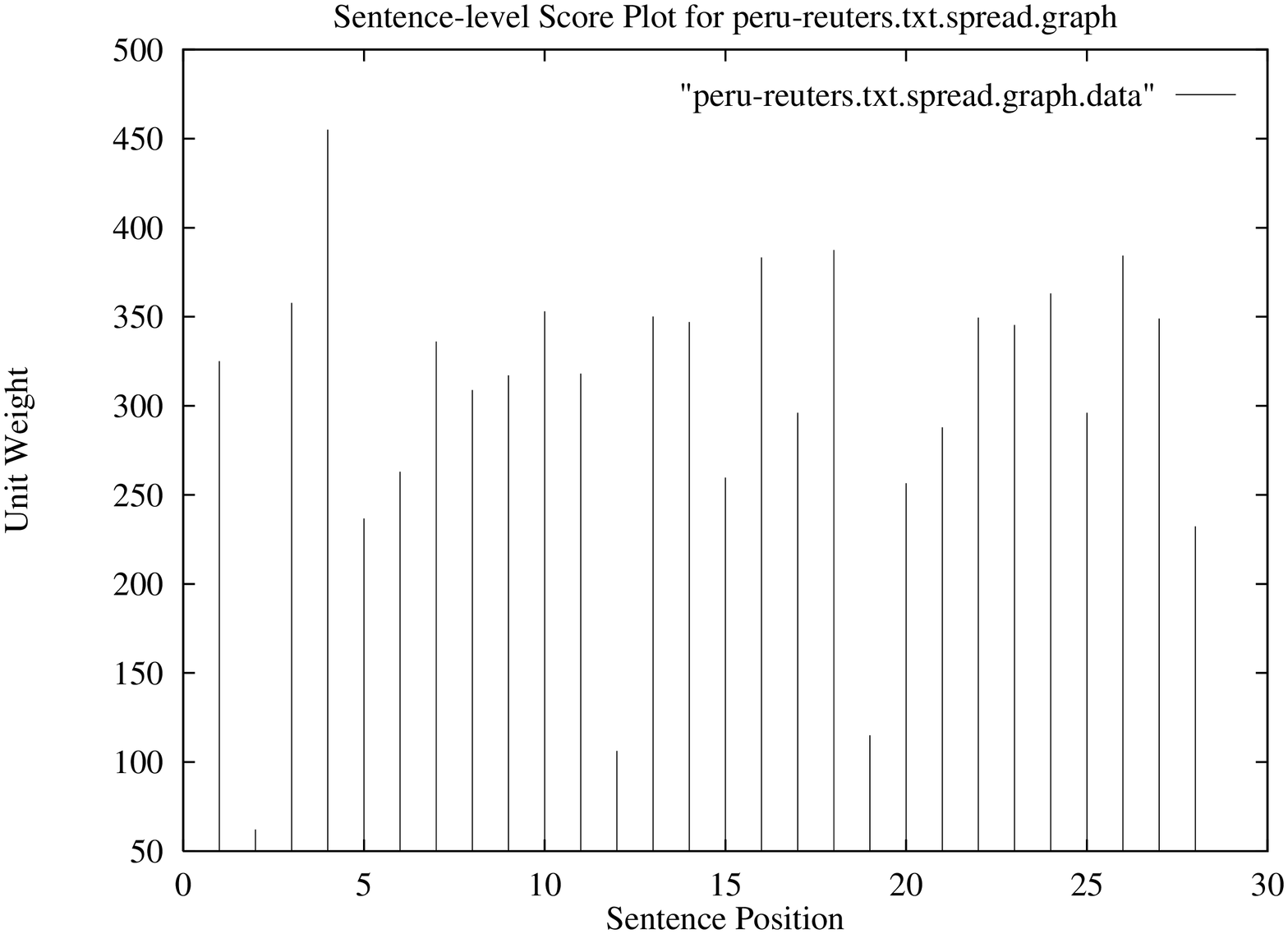,height=1.4in,angle=270}}
\caption{Activation Weights from Graph after Spreading Activation (Reuters news; topic: {\it Tupac Amaru})}
\label{fig-4b}
\end{figure}}

Unlike much previous use of spreading activation methods for query
expansion, as a part of information retrieval \cite{Salton-2}
\cite{Arizona}, our use of spreading activation is to reweight the
words in the document rather than to decide for each word whether it
should be included or not. The later synthesis module determines the
ultimate selection of nodes based on node weight as well as its
relationship to other nodes. As a result, we partially insulate the
summary from the potential sensitivity of the spreading to the choice
of starting nodes and search extent. For example, we would get the
same results for {\it Tupac Amaru} as the topic as with {\it
MRTA}. Further, this means the spreader need not capture all nodes
that are relevant to a summary directly, but only to suggest new
regions of the input text that may not immediately appear to be
related. 

This has distinct advantages compared to certain information retrieval
methods which simply find regions of the text similar to the query.
For example, the Reuters sentence 4 plotted in Figure~\ref{fig-4b} and
shown in Figure~\ref{fig-5} might have been found via an information
retrieval method which matched on the query {\it Tupac Amaru}
(allowing for {\it MRTA} as an abbreviated alias for the
name). However, it would have not found other information related to
the {\it Tupac Amaru}: In the Reuters article, the spreading method
follows a link from {\it Tupac Amaru} to {\it release} in sentence 4
(via ADJ), to other instances of {\it release} via the SAME link,
eventually reaching sentence 13 where {\it release} is ADJ to the name
{\it Victor Polay} (the group's leader).  Likewise, the algorithm
spreads to sentences 26 and 27 in that article which mention {\it
MRTA} but not {\it Tupac Amaru}.  In the AP article, a thesaurus link
becomes more useful in establishing a similar connection: it is able
to find a direct link from {\it Tupac Amaru} to {\it leaders} (via
ADJ) in sentence 28, and from there to its synonym {\it chief} in
sentence 29 (via ALPHA), which is ADJ to {\it Victor Polay}\footnote
{Of course, the relation could also be found if the system could
correctly interpret the expressions {\it its chief} in the AP article
and {\it their leader} in the Reuters article.}.

\section {Summarizing Multiple Documents by Graph Matching}

The goal of FSD-Graphs is to find the concepts which best describe the
similarities and differences in the given regions of text. It does
this by first finding which concepts (nodes) are common and which are
different. The computation of common nodes given graphs G1 and G2 is
given by $Common = \{ c | concept\_match(c, G1) \& concept\_match(c,
G2) \}$. Differences are computed by: $Differences = (G1 \cup G2) -
Common$.  $concept\_match(c, G)$ holds if there is a c1 in G such that
either $word(c1) =_{stem} word(c)$, or $synonym(word(c1), word(c))$. The
user may provide a threshold on the minimal number of uniquely covered
concepts, or on the minimal coverage weight. 

{\small \begin{figure}
\centerline{\psfig{figure=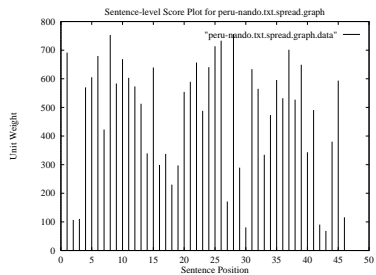,height=1.4in,angle=270}}
\caption{Activation Weights from Spread Graph 
(AP news; topic: {\it Tupac Amaru})}
\label{fig-4c}
\end{figure}}

Currently, the synthesis module simply outputs the set of sentences
covering the shared terms and the set of sentences covering the unique
terms, hilighting the shared and unique terms in each, and indicating
which document the sentence came from. This is something of a fallback
arrangement, as the abstraction built is not represented to the user.
In the next phase of research, we expect to better exploit the
concepts in the text, their semantic relations, and concepts from the
thesaurus to link extracts into abstracts.

Sentence selection is based on the coverage of nodes in the common and
different lists. Sentences are greedily selected based on the average
activated weight of the covered words: For a sentence s, its score in
terms of coverage of common nodes is given by $score(s) =
\frac{1}{|c(s)|}\sum_{i=1}^{|c(s)|}weight(w_{i})$, where $c(s) = \{ w
| w \epsilon Common \cap s \}$. The score for Differences is
similar. The user may specify the maximal number of sentences in a
particular category (common or different) to control which sentences
are output.

As an example, consider the application of FSD-Graphs to the activated
graph in Figure~\ref{fig-4b} (the Reuters article) and an activated
graph in Figure~\ref{fig-4c} (an AP article of the same date
describing the same hostage crisis). The activated graphs had 94 words
in Common, out of 343 words for the former graph and 414 for the
latter. The algorithm extracts 37 commonalities, with the
commonalities with the strongest associations being on top.  The high
scoring commonalities and differences are the ones shown in
Figure~\ref{fig-5}.  The algorithm discovers that both articles talk
about {\it Victor Polay} (e.g., the Reuters sentence 13 mentioned
earlier, and the AP sentence 29), {\it Fujimori}, {\it Japanese
ambassador}, {\it residence}, and {\it cabinet}. Notice that the
system is able to extract commonalities without {\it Tupac Amaru}
being directly present. Regarding differences, the algorithm discovers
that the AP article is the only one to explain how the rebels posed as
waiters (sentence 12) and the Reuters article is the only one which
told how the rebels once had public sympathy (sentence 27).

{\small
\begin {table*}[t]
\centering
\begin{tabular}{|c|c|c|}\hline
     {\bf Metric \/}  &                             {\bf Full-Text \/} &                               {\bf Summary \/} \\ \hline
Accuracy (Precision, Recall)     &       30.25, 41.25 & 25.75, 48.75    \\ \hline
Time (mins)       &   {\bf 24.65 \/}    &  {\bf 21.65 \/} \\ \hline
Usefulness of text in deciding relevance (0 to 1)       &   .7    &  .8 \\ \hline
Usefulness of text in deciding irrelevance (0 to 1)       &   .7    &  .6 \\ \hline
Preference for more or less text    & ``Too Much Text.''       & ``Just Right.''  \\ \hline
\end{tabular}
\caption{Summaries versus Full-Text: Task Accuracy, Time, and User Feedback}
\label{table-1}
\end{table*}
}

{\small
\begin {table*}[t]
\centering
\begin{tabular}{|c|c|}\hline
     {\bf Condition \/}  &                             {\bf Without Subgraph Extraction \/} \\ \hline
{\bf Without Spreading \/}    &    4.6, 1.7 \\ \hline
{\bf With Spreading \/}       &    5.6, 3.9 \\ \hline
\end{tabular}
\caption{Mean Ratings of Multi-Document Summaries (Commonalities, Differences)}
\label{table-2}
\end{table*}
}

\section {Evaluation}

\subsection {Effectiveness of Spreading Activation Graph Search}

Methods for evaluating text summarization approaches can broadly
classified into two categories. The first is an extrinsic evaluation
in which the quality of the summary is judged based on how it effects
the completion of some other task. The second approach, an intrinsic
evaluation, judges the quality of the summarization directly based on
user judgements of informativeness, coverage etc. In our
evaluation we performed both type of experiments.

In our extrinsic evaluation we evaluated the usefulness of
Graph-Search (spreading) in the context of an information retrieval
task. In this experiment, subjects were informed only that they were
involved in a timed information retrieval research experiment. In each
run, a subject was presented with a pair of query and document, and
asked to determine whether the document was relevant or irrelevant to
the query. In one experimental condition the document shown was the
full text, in the other the document shown was a summary generated
with the top 5 sentences. Subjects (four altogether) were rotated
across experimental conditions, but no subject was in both conditions
for the same query-document pair. We hypothesized that if the
summarization was useful, it would result in savings in time, without
significant loss in accuracy.

Four queries, were preselected from the TREC \cite{Harman} collection
of topics, with the idea of exploiting their associated (binary)
relevance judgments.  These were 204 (``Where are the nuclear power
plants in the U.S.  and what has been their rate of production?''),
207 (``What are the prospects of the Quebec separatists achieving
independence from the rest of Canada?''), 210 (``How widespread is the
illegal disposal of medical waste in the U.S. and what is being done
to combat this dumping?''), and 215 (``Why is the infant mortality
rate in the United States higher than it is in most other
industrialized nations?'')\footnote{Given a TREC query and a document
to be summarized, the entry nodes for spreading activation are those
document nodes which are $stem_{=}$ to non-stop-word terms found in
the TREC query.}.

A subset of the TREC collection of documents was indexed
using the SMART retrieval system from Cornell \cite{Buckley}. Using
SMART, the top 75 hits from each query was reserved for the
experiment. Overall, each subject was presented with four batches of
75 query-document pairs (i.e., 300 documents were presented to each
subject), with a questionnaire after each batch.  Accuracy metrics in
information retrieval include precision (percentage of retrieved
documents that are relevant, i.e., number retrieved which were
relevant$/$total number retrieved) and recall (percentage of relevant
documents that are retrieved, i.e., number retrieved which were
relevant$/$total number known to be relevant).

In Table~\ref{table-1}, we show the average precision and average
recall over all queries (1200 relevance decisions altogether). The
table shows that when the summaries were used, the performance was
faster than with full-text (F=32.36, p $<$ 0.05, using analysis of
variance F-test) without significant loss of accuracy. While we would
expect shorter texts to take less time to read, it is striking that
these short extracts (on average, one seventh of the length of the
corresponding full-text - which in turn was on average about 200
words long) are effective enough to support accurate retrieval.  In
addition, the subjects' feedback from the questionnaire (shown in the
last three rows of the table) indicate that the spreading-based
summaries were found to be useful.

\subsection {Effectiveness of FSD-Graphs}

We also performed an intrinsic evaluation of our summarization
approach by generating summaries from FSD-graphs with and without
spreading activation.  In this evaluation we used user judgements to
assess directly the quality of FSD-Graphs using spreading to find
commonalities and differences between pairs of documents. When
FSD-Graphs is applied to ``raw'' graphs which are not reweighted by
spreading, the approach does not exploit at all the relational model
of summarization. We hypothesized that the spreading or
Extract-Subgraphs methods would result in more pertinent summaries
than with the ``raw'' graphs. For this experiment, 15 pairs of
articles on international events were selected from searches on the
World Wide Web, including articles from Reuters, Associated Press, the
Washington Post, and the New York Times. 

\onecolumn

{\small \begin{figure} 
\psfig{figure=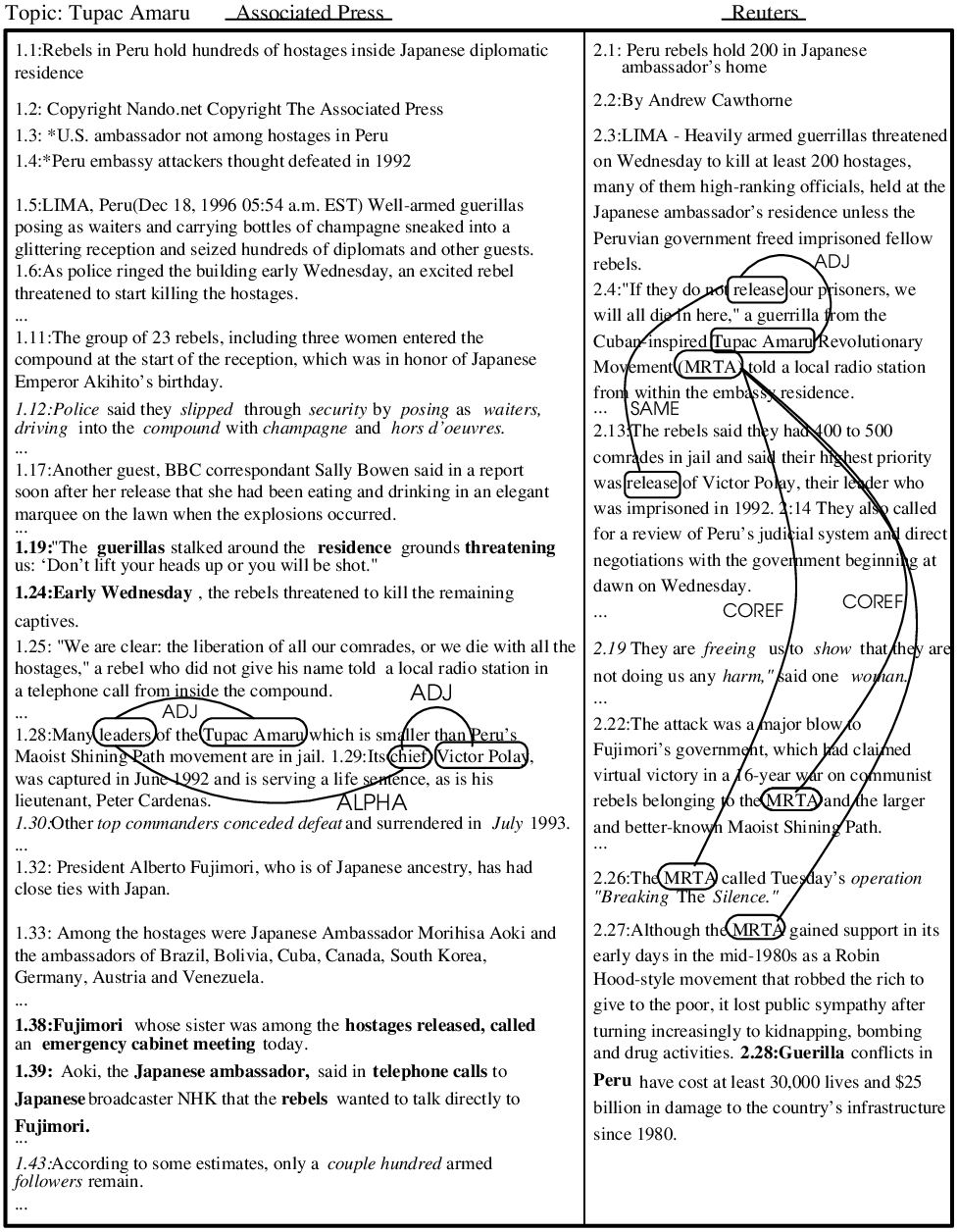}
\caption{Texts of two related articles. The top 5 salient sentences
containing common words have these common words in bold face;
likewise, the top 5 salient sentences containing unique words have
these unique words in italics. }
\label{fig-5}
\end{figure}}

\twocolumn 

Pairs were selected such that each member of a pair was closely
related to the other, but by no means identical; the pairs were drawn
from different geopolitical regions so that no pair was similar to
another. The articles we found by this method happened to be short
ones, on average less than two hundred words long.  A distinct topic
was selected for each pair, based on the common activators
method. Summaries were then generated both with no spreading using
only the raw tf.idf weights of the words, and with spreading.  Three
subjects were selected, and each subject was presented with a series
of Web forms. In each form, the subject was shown a pair of articles,
along with a summary of their similarities and a summary of their
differences, with respect to the pair topic. Each subject was asked to
judge on a scale of 1 (bad) to 10 (good) how well the summaries
pinpointed the similarities and differences with respect to the
topic. Each subject was rotated at random through all the forms and
experimental conditions, so that each subject saw 60 different forms
and made 120 decisions (360 data points altogether).

As shown in Table~\ref{table-2}, using spreading results in improved
summaries over not using spreading for both commonalities and
differences. It is interesting to note that the biggest improvement
comes from the differences found using spreading. This reflects the
fact that the spreading algorithm uses the topic to constrain and
order the differences found.  By contrast, in a tf.idf weighting
scheme, words which are globally unique are rewarded highest
regardless of their link to the topic at hand. 

\section{Conclusion}

We have described a new method for multi-document summarization based
on a graph representation for text.  The summarization exploits the
results of recent progress in information extraction to represent
salient units of text and their relationships.  By exploiting {\it
relations} between units and the {\it perspective} from which the
comparison is desired, the summarizer can pinpoint similarities and
differences. Our approach is highly domain-independent, even though we
have illustrated its power mainly for news articles. Currently, the
synthesis component is rudimentary, relying on sentence extraction to
exemplify similarities and differences. In future work, we expect to
more fully exploit $alpha$ links, especially by more systematic
extraction of semantic distance measures (along with corpus-based
statistics) from WordNet. We also plan to exploit both text and
thesaurus concepts to link extracts into abstracts.

\section* {Acknowledgments}

We are grateful to the following colleagues: Gary Klein for help with
experimental design and evaluation, David House for extensive work on
user interfaces, and Barbara Gates for SMART support.

\end{document}